\documentclass[10pt,twocolumn,letterpaper]{article}

\usepackage{cvpr}              
\usepackage{graphicx}
\definecolor{cvprblue}{rgb}{0.21,0.49,0.74}
\usepackage[pagebackref,breaklinks,colorlinks,allcolors=cvprblue]{hyperref}

\def\paperID{23} 
\def\confName{MMRAgI Workshop CVPR}
\def\confYear{2026}

\title{Inference-Time Agentic Decision Rules Beat Longer Evolving Search\\
for Multi-Image Medical Reasoning}

\author{
Site Li \qquad Jianyi Hao \qquad Xiaofeng Liu \\
Yale University \\
}

\begin{document}
\maketitle

{\let\thefootnote\relax\footnotetext{Presented at the CVPR 2026 Workshop on Multi-Modal Reasoning for Agentic Intelligence.}}

\begin{abstract}
Multi-image medical VQA is not merely a prompt-length problem; it is a fundamental challenge of agentic decision-making. Medical vision-language agents must aggregate evidence across ordered images, remain robust to answer-order perturbations, and avoid overfitting to noisy search-time feedback. We study MedFrameQA through a controlled comparison of five inference-time agentic strategies, optimized using the same high-budget ShinkaEvolve configuration and evaluated on a reproducible internal frozen split (1,331 evolution, 665 holdout, 855 final test). Across five independent repeated runs, the strongest method emerges as the simplest robust aggregator: the \textbf{order-vote} policy achieves $57.89 \pm 0.65\%$ final-test accuracy, significantly outperforming the fixed baseline ($52.73 \pm 0.42\%$) and the more complex, albeit brittle, order-rerank variant ($55.79 \pm 0.43\%$). Paired bootstrap analysis confirms these significant gains. Extending the evolutionary search budget from 50 to 100 generations yields no generalization benefit: while holdout performance marginally increases, final-test accuracy drops from $57.89\%$ to $56.02\%$. Our findings suggest that for multi-image medical reasoning, defining the correct agentic decision rule is substantially more impactful than expanding the optimization search budget. 
\end{abstract}

\section{Introduction}

\begin{figure}[t]
    \centering
    \includegraphics[width=\linewidth]{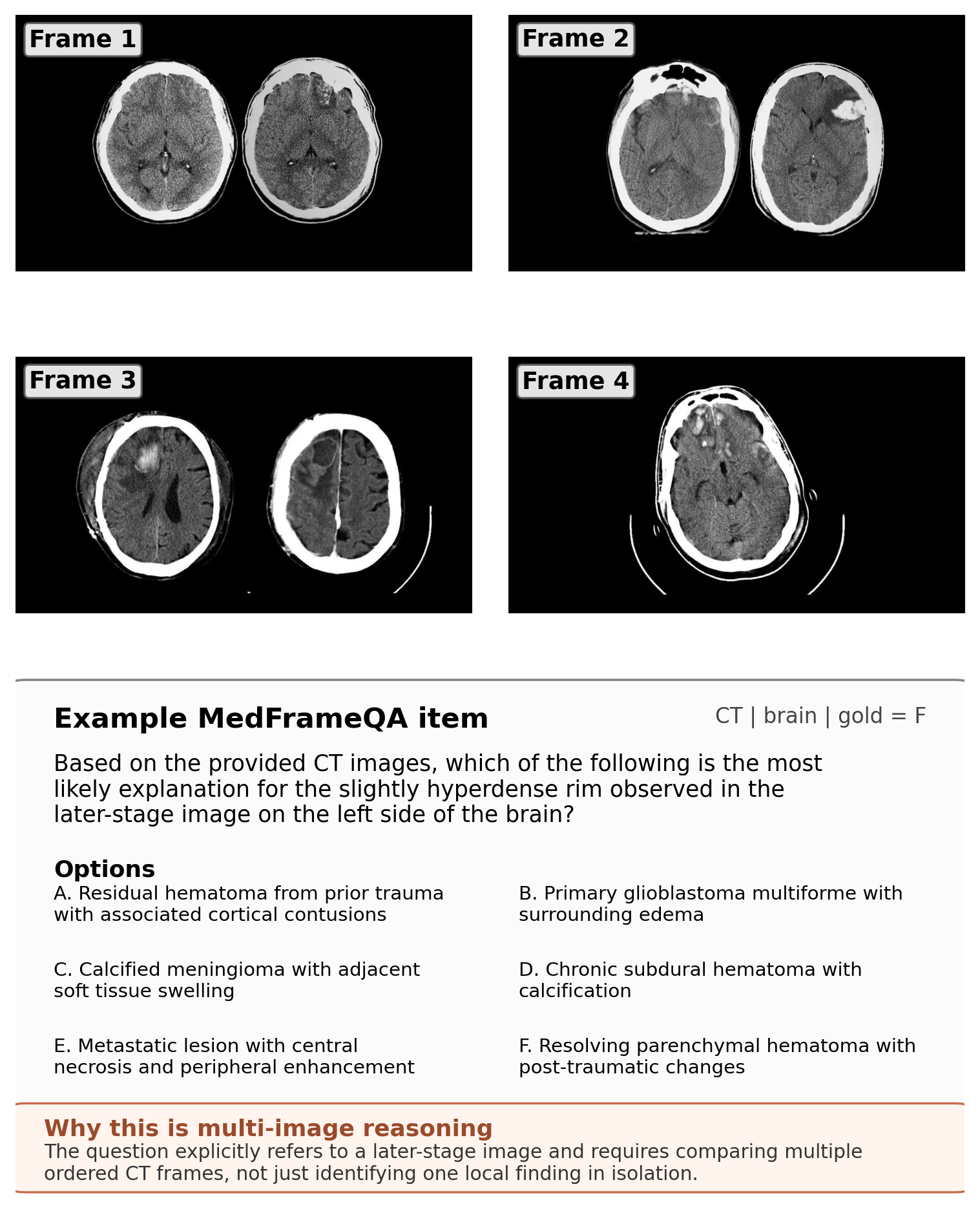}
    \caption{A real MedFrameQA example optimized for single-column readability. The task requires comparing evidence across an ordered image sequence rather than answering from one frame in isolation.}
    \label{fig:task-example}
\end{figure}

Medical vision-language modeling is moving beyond single-image question answering toward tasks that require reasoning over \emph{ordered sets} of images. This shift is particularly critical for clinical use cases, where longitudinal scans, multi-view studies, and structured follow-ups are routine. In this regime, an AI system functions less as a static classifier and more as an \emph{investigative agent}: it must integrate evidence across a sequence where the clinical meaning of a finding can dynamically change from one frame to the next. Recent multi-image and interleaved benchmarks make this challenge increasingly explicit, both in medical domains and general multimodal evaluation \citep{yu2025medframeqa,jiang2024mantis,meng2025mmiu,wang2025muirbench,xia2025mmie}.

This setting exposes a critical limitation in how prompt optimization is traditionally framed. While extended context windows, multimodal chain-of-thought scaffolds, and many-shot prompting can improve reasoning behaviors in certain environments \citep{zhang2023mmcot,jiang2024manyshot}, they do not inherently define an agent's policy for aggregating evidence, mitigating answer-order sensitivity, or triggering secondary arbitration. For ordered multi-image medical VQA, the central bottleneck is not simply prompt wording; it is the underlying \emph{inference-time agentic decision rule}.

Prompt-program evolution offers a natural way to optimize such systems without gradient updates. Recent work has shown that prompts and lightweight programs can be searched, refined, or evolved through automatic prompt engineering, textual gradients, optimizer-style prompting, reinforcement learning, and self-referential mutation \citep{zhou2023ape,pryzant2023protegi,yang2024opro,deng2022rlprompt,guo2023evoprompt,fernando2023promptbreeder,khattab2024dspy,yuksekgonul2024textgrad,kwon2024stableprompt}. ShinkaEvolve extends this line by treating prompts and code-like artifacts as evolvable programs \citep{lange2025shinkaevolve}. In this paper, however, ShinkaEvolve is not the endpoint; it is the shared search engine that lets us compare inference-time strategies under one controlled budget. Once the setting becomes genuinely multimodal and sequence-sensitive, the more consequential question is: \emph{what kind of agentic decision rule is worth evolving at all?}

That distinction matters because superficially similar prompting methods implement fundamentally different reasoning policies. A fixed all-images prompt, a textual reasoning scaffold, a multi-view voting program, and a reranking-heavy program all consume the same multimodal inputs, yet they impose distinct decision boundaries. Some act as one-shot reactive selectors; others function as structured agents that aggregate multiple local decisions or execute a conditional second arbitration stage when early evidence conflicts. In our experiments, this choice of decision-rule architecture dictates the performance ceiling, dominating the marginal gains obtainable from an extended evolutionary budget.

This observation motivates the core question of the paper: \emph{when the benchmark requires ordered multi-image reasoning, which inference-time agentic strategy actually survives a strict holdout-to-final validation pipeline?} Explicit reasoning scaffolds might help because they encourage structured inspection. Reranking might help because it compares plausible options more carefully. Or answer-order sensitivity might be the dominant nuisance factor, in which case robust vote aggregation should matter more than extra reasoning text. Our experiments are designed to adjudicate among these alternatives under one controlled protocol.

Our central claim is simple: \textbf{decision-rule choice dominates additional search budget}. Under a protocol that cleanly separates search, holdout selection, and independent final testing, the winning method is neither the most verbose strategy nor the deepest arbitration stack. It is a simpler order-robust voting strategy whose inductive bias is better matched to multi-image decision making. This is exactly why the appendix budget-sensitivity result matters: if the remaining bottleneck were insufficient search depth, then 100 generations should help. They do not.

Concretely, the paper makes three claims:
\begin{itemize}[leftmargin=1.5em]
    \item We provide a controlled comparison of five inference-time agentic strategies for multi-image medical VQA under the same high-budget ShinkaEvolve setup and the same internal frozen split.
    \item We show that a lightweight \textbf{order-vote} policy consistently outperforms both a stable fixed baseline and a more expensive order-rerank variant on the final test set.
    \item We show that extending the search budget from 50 to 100 generations does not improve final generalization, indicating that longer search can amplify selection noise rather than produce a better final program.
\end{itemize}

This framing is timely for medical multimodal evaluation. Recent medical VLMs and Med-LVLMs have rapidly expanded scale and coverage \citep{moor2023medflamingo,li2023llavamed,zhang2023biomedgpt,hu2024omnimedvqa,chen2024chexagent,lin2025healthgpt}, but careful evaluations still show that domain adaptation and instruction following do not automatically produce robust clinical multimodal reasoning \citep{jeong2024medicaladapt,agrawal2025evaluationillusion}. Our results suggest that for ordered multi-image reasoning, the design of the agent's decision rule deserves at least as much attention as prompt verbosity or model scale.

\section{Related Work}

\begin{figure*}[t]
    \centering
    \includegraphics[width=\textwidth]{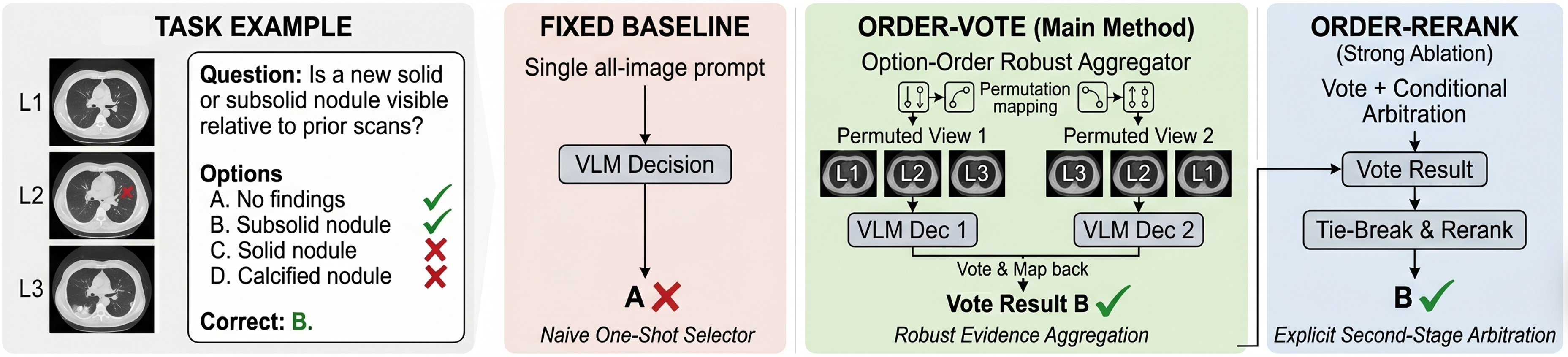}
    \caption{Inference-time control structures for the three most important agentic strategies. The crucial difference is not prompt length but the decision rule: one-shot selection (Fixed), mapped voting over reordered answer views (Order-Vote), and vote followed by conditional second-stage comparison (Order-Rerank).}
    \label{fig:method-overview}
\end{figure*}

\noindent\textbf{Medical vision-language modeling.}
General-purpose VLM development has rapidly improved the quality of instruction-following multimodal systems, with representative milestones including Flamingo, BLIP-2, InstructBLIP, LLaVA, and Qwen-VL \citep{alayrac2022flamingo,li2023blip2,dai2023instructblip,liu2023llava,bai2023qwenvl}. Medical adaptations and specialist models have followed quickly, including Med-Flamingo, LLaVA-Med, BiomedGPT, OmniMedVQA, CheXagent, and HealthGPT \citep{moor2023medflamingo,li2023llavamed,zhang2023biomedgpt,hu2024omnimedvqa,chen2024chexagent,lin2025healthgpt}. Our goal is different: we do not propose a new medical base model, but instead study how inference-time prompt programs should be structured when the benchmark requires an agent to reason over an \emph{ordered set} of images.

\noindent\textbf{Medical VQA and benchmark design.}
Medical VQA benchmarks span multiple task styles, from early radiology and pathology datasets such as VQA-RAD, PathVQA, and SLAKE to broader medical instruction-tuning and evaluation resources such as PMC-VQA, MLeVLM, and OmniMedVQA \citep{lau2018vqarad,he2020pathvqa,liu2021slake,zhang2023pmcvqa,xu2024mlevlm,hu2024omnimedvqa}. MedFrameQA is especially relevant here because it was explicitly designed for \emph{multi-image} clinical reasoning, with each question grounded in two to five images rather than a single frame \citep{yu2025medframeqa}. This makes it a useful testbed for studying whether agentic program search should emphasize fixed context assembly, stepwise scaffolds, or more structured decision rules such as voting and reranking.

\noindent\textbf{Multi-image and multimodal reasoning benchmarks.}
Recent multimodal benchmarks increasingly test reasoning rather than only local recognition, including MME, SEED-Bench, MMMU, and multi-image/interleaved benchmarks such as MANTIS, MMIU, MuirBench, and MMIE \citep{fu2023mme,li2023seedbench,yue2023mmmu,jiang2024mantis,meng2025mmiu,wang2025muirbench,xia2025mmie}. These datasets help motivate our framing: once questions depend on multiple images, temporally ordered evidence, or interleaved context, the relevant comparison is no longer only prompt phrasing but also the underlying agentic decision rule.

\noindent\textbf{Multimodal Agentic Reasoning.}
The emergence of Multimodal Foundation Models (MFMs) has driven the development of agentic systems capable of interacting with complex environments, spanning from software navigation (e.g., OS-Copilot) to autonomous scientific research. While much of the agentic literature focuses on embodied interaction or tool use, clinical reasoning agents face a distinct bottleneck: cross-modal semantic understanding and robust evidence aggregation over time. Our work explores the decision policies that empower these agents to resolve ambiguities across an ordered sequence of multimodal inputs, directly addressing the core focus of enhancing agentic multimodal perception and reasoning.

\noindent\textbf{Prompt and program evolution.}
Prompt and program optimization now spans automatic prompt engineering, textual-gradient methods, optimizer-style prompting, reinforcement learning, declarative LM program compilation, and evolutionary mutation \citep{zhou2023ape,pryzant2023protegi,yang2024opro,deng2022rlprompt,guo2023evoprompt,fernando2023promptbreeder,khattab2024dspy,yuksekgonul2024textgrad,kwon2024stableprompt}. Program evolution frameworks such as ShinkaEvolve extend this line by treating prompts and code-like artifacts as evolvable programs under explicit search-time feedback \citep{lange2025shinkaevolve}. Our empirical angle is complementary: we hold the high-budget ShinkaEvolve engine fixed and ask which \emph{agentic strategy} benefits most from search in a multi-image medical setting.

This emphasis clearly differentiates our contribution from a generic prompt-tuning paper. The empirical object of comparison is not merely a surface string but rather a structured inference policy that determines how an agent groups images, perturbs the options to check robustness, and decides when additional deliberative comparison should be triggered.

\section{Task Setup and Internal Frozen Split}

We evaluate on MedFrameQA \citep{yu2025medframeqa}, a benchmark built for multi-image medical VQA with clinical reasoning over ordered image sets. This positions it differently from earlier medical VQA resources such as VQA-RAD, PathVQA, and SLAKE, which are valuable but largely single-image in structure \citep{lau2018vqarad,he2020pathvqa,liu2021slake}. Each example contains between two and five images drawn from multiple modalities, together with a multiple-choice question whose answer requires reasoning over the full sequence rather than a single frame.

\begin{table}[t]
\centering
\resizebox{\linewidth}{!}{%
\begin{tabular}{lp{1.3cm}p{5.2cm}}
\toprule
Split & Size & Purpose \\
\midrule
Evolution pool & 1,331 & Longer evaluation pool used for milestone reevaluation and evolutionary search \\
Selection holdout & 665 & Model-selection split for choosing the final generation within each run \\
Independent final test & 855 & Frozen evaluation split used only for final reporting and post-hoc analysis \\
\bottomrule
\end{tabular}%
}
\caption{Internal frozen split used throughout the paper. All methods share the same group-preserving split by \texttt{video\_id}.}
\label{tab:split-summary}
\end{table}

\begin{figure*}[t!]
    \centering
    \includegraphics[width=1\linewidth]{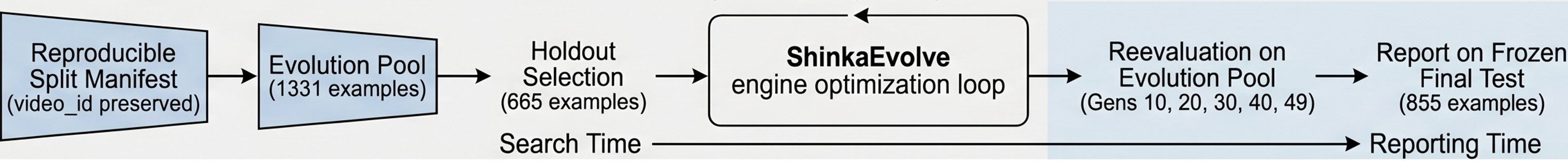}
\caption{End-to-end protocol. Search uses a fixed \texttt{search\_mini} subset for fast feedback, but generation selection always uses the same holdout split and reporting always uses the same independent final test.}
    \label{fig:protocol-overview}
\end{figure*}

Figure~\ref{fig:task-example} illustrates the benchmark format we target in this paper. The images are not interchangeable: later frames can refine or even change the clinical interpretation suggested by earlier ones. This is exactly the regime where naive prompt extensions are often insufficient, because the inference-time program must act as a sequential observer, deciding how to integrate evidence across the full sequence and adjudicate among multiple answer candidates.

This benchmark structure also sharpens the methodological stakes. In MedFrameQA, the critical failure modes are incomplete evidence aggregation, spurious reliance on one frame, and sensitivity to answer order. Those are not naturally solved by more text alone; they require a better agentic inference policy.

The public release used in our environment exposes one visible test split. To avoid using the same public pool for both search and final reporting, we construct a reproducible \emph{internal frozen split}. The split is group-preserving by \texttt{video\_id}, so all examples derived from the same underlying case stay in the same partition. We then optimize the partition assignment with a multi-start bundle-exchange search that balances modality and answer-label distributions. The resulting split contains 2,851 examples in total:
\begin{itemize}[leftmargin=1.5em]
    \item \textbf{Evolution pool}: 1,331 examples
    \item \textbf{Selection holdout}: 665 examples
    \item \textbf{Independent final test}: 855 examples
\end{itemize}

This split is frozen and versioned in our codebase as \texttt{medframeqa\_split\_manifest\_v2}. All methods share the exact same split, evaluation protocol, and downstream post-hoc validation pipeline. It is an \emph{internal frozen split}, not an official hidden test set, and all claims in this paper are made under that documented protocol.

Two details matter for interpretation. First, the split avoids case leakage by keeping each \texttt{video\_id} in exactly one partition. Second, the split optimizer balances modality and answer distributions, making final comparisons more interpretable than a naive random partition of the visible public pool.

The split protocol therefore plays a substantive role rather than merely an implementation role. If search, model selection, and final reporting all used the same visible pool, the main empirical question would be confounded by selection bias. By separating evolution, holdout selection, and final testing on a frozen case-preserving split, we can tell whether a method is genuinely stronger or simply better at fitting the intermediate selection signal.

\section{Inference-Time Agentic Strategies}

We compare five inference-time agentic strategies. Each strategy is implemented as a restricted prompt-program configuration whose content can be evolved by ShinkaEvolve, while the surrounding runtime and evaluation logic remain fixed.

\begin{table}[t]
\centering
\small
\resizebox{\linewidth}{!}{{\begin{tabular}{p{1.7cm}p{3.0cm}p{3.1cm}}
\toprule
Strategy & Core idea & Inference rule \\
\midrule
Fixed & Single all-image prompt & Reactive one-shot decision \\\midrule
Reasoning & Structured inspection scaffold & Reflective one-shot decision \\\midrule
Order-Vote & Option-order perturbation & Robust aggregation vote \\\midrule
Order-Rerank & Vote + pairwise correction & Deliberative reranking \\\midrule
Order-Vote+ & Vote + selective top-2 check & Conditional second stage \\
\bottomrule
\end{tabular}}}
\caption{Summary of the five agentic strategies. The key distinction is not just prompt wording, but the decision procedure applied over multi-image inputs and answer options.}
\label{tab:method-families}
\end{table}

\begin{figure*}[t]
    \centering
    \includegraphics[width=\textwidth]{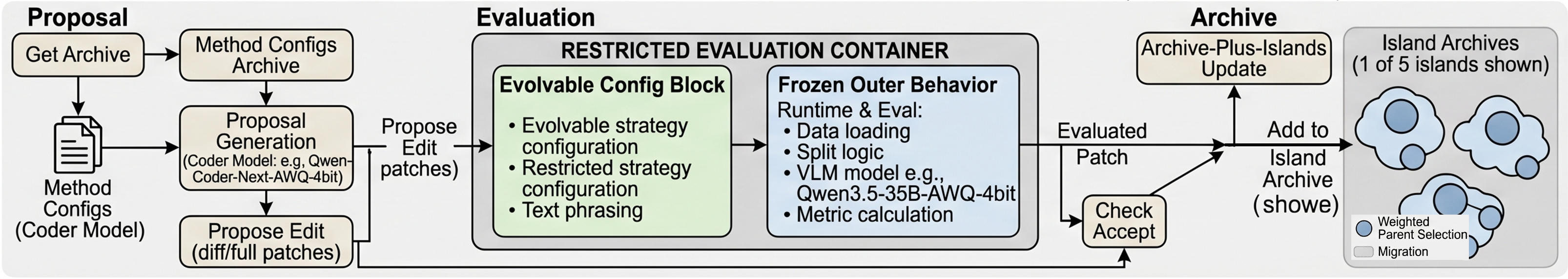} 
\caption{Zoom-in of the intra-run search loop. Evolution edits only the strategy-specific config block, scores candidates on \texttt{search\_mini}, keeps promising variants in an archive-plus-islands loop, periodically reevaluates them on the larger evolution pool, and only then performs holdout selection and final-test reporting.} 
    \label{fig:evolution-process}
\end{figure*}

\paragraph{Fixed.}
The fixed strategy is the minimal baseline. It presents the full ordered image set, the question, and the answer options once, and asks the model to output a single letter. This tests whether prompt evolution alone can improve a straightforward, reactive all-images decision policy.

\paragraph{Reasoning.}
The reasoning strategy adds a structured scaffold that instructs the model to inspect each image, relate local findings to candidate options, and prefer answers that explain the full sequence. This isolates the effect of more explicit stepwise prompting without changing the underlying one-shot decision rule.

\paragraph{Order-Vote.}
The order-vote strategy perturbs the option order across multiple local views, maps local predictions back to the original label space, and aggregates them by vote. Its purpose is to reduce answer-order sensitivity while preserving a computationally efficient inference-time procedure, a crucial requirement for responsive multimodal agents. This is our strongest method.

\paragraph{Order-Rerank.}
Order-rerank starts from the same reordered local views as order-vote but adds an extra pairwise reranking stage when disagreements remain. It is therefore more computationally expensive and more brittle, but tests whether explicit second-stage deliberative comparison is better than simple vote aggregation.

\paragraph{Order-Vote+.}
Order-vote+ is a selective top-2 correction variant. It only triggers an additional comparison when the first-stage vote is uncertain. This strategy is kept as an appendix exploration rather than a main-table method because it is currently supported by only one complete repeat and does not outperform order-rerank on final test.

Figure~\ref{fig:method-overview} and Table~\ref{tab:method-families} make clear why we treat these as different \emph{strategies} rather than trivial prompt variants. Fixed and reasoning rely on a single final decision after a single forward pass over the ordered image set. Order-vote changes the decision rule by aggregating predictions over answer-order perturbations. Order-rerank further introduces an explicit second stage whose complexity is conditional on disagreement. These changes alter both search behavior and evaluation cost, which is why our comparison is framed around inference-time strategy rather than prompt text alone.

\section{Experiments and Results}

\begin{table}[t]
\centering
\small
\resizebox{\linewidth}{!}{\begin{tabular}{lcccc}
\toprule
Method & Runs & Holdout (\%) & Final (\%) & Invalid (\%) \\
\midrule
Fixed & 5 & 53.50 $\pm$ 0.78 & 52.73 $\pm$ 0.42 & 0.80 \\
Order-Vote & 5 & 62.50 $\pm$ 1.08 & 57.89 $\pm$ 0.65 & 3.20 \\
Order-Rerank & 5 & 59.52 $\pm$ 0.36 & 55.79 $\pm$ 0.43 & 10.40 \\
\bottomrule
\end{tabular}}
\caption{Main repeat-only high-budget results. All numbers come from the publication-clean repeated runs under the same protocol.}
\label{tab:main-results}
\end{table}

\paragraph{Evolution engine.}
All prompt-program strategies are evolved with ShinkaEvolve \citep{lange2025shinkaevolve}. We use a high-budget local setup with 5 islands, a 40-entry archive, weighted parent selection, migration every 10 generations, and up to 3 patch attempts per generation. Patch types are restricted to \texttt{diff} and \texttt{full}; we do not use cross-over style edits because our evolvable space is primarily configuration-like and cross-style edits were empirically destabilizing. In this paper, the value of ShinkaEvolve is diagnostic as much as optimizational: the shared search engine reveals which decision rules benefit from search and which ones do not.

Figure~\ref{fig:evolution-process} deliberately complements rather than repeats Figure~\ref{fig:protocol-overview}. Figure~\ref{fig:protocol-overview} gives the end-to-end protocol; Figure~\ref{fig:evolution-process} isolates the intra-run search loop. The system is not evolving arbitrary runtime code. It is evolving restricted strategy-specific configuration blocks under a shared proposal, evaluation, archive, migration, and holdout-selection pipeline.

\begin{figure*}[t]
    \centering
    \includegraphics[width=1\linewidth]{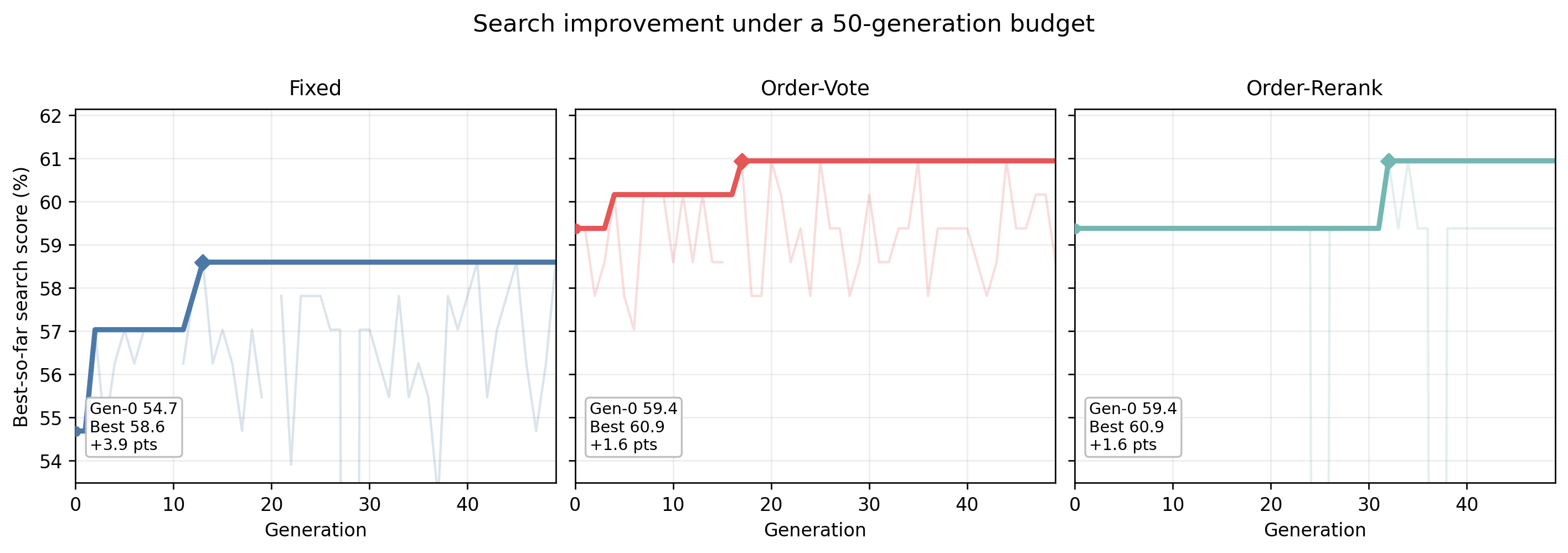}
    \caption{Representative 50-generation search progress. Each panel overlays the noisy raw search trace with a bold best-so-far curve, making within-budget gains visible without letting invalid generations dominate the visual story.}
    \label{fig:evolution-progress}
\end{figure*}

\begin{table}[t!]
\centering
\small
\begin{tabular}{lccccp{2.7cm}}
\toprule
Method & Runs & Holdout (\%) & Final (\%) & Invalid (\%)   \\
\midrule
Reasoning & 1 & 53.98 & 51.58 & 0.00  \\
Order-Vote+ & 1 & 54.89 & 55.32 & 8.00   \\
\bottomrule
\end{tabular} 
\caption{Single-run weak baselines and exploratory variants. They are reported for completeness but are not used in the main table decision.}
\label{tab:weak-baselines}
\end{table}

\noindent\textbf{Models and serving.}
Proposal generation uses a locally served coder model (\texttt{Qwen3-Coder-Next-AWQ-4bit}) through an OpenAI-compatible endpoint. Evaluation uses a single shared local VLM endpoint backed by \texttt{Qwen3.5-35B-A3B-AWQ-4bit}. Embedding support for ShinkaEvolve uses \texttt{BAAI/bge-m3}. All serving is local and cost-free in API terms, but runtime cost is still reflected in VLM call counts and wall-clock time.

\noindent\textbf{Budget and selection.}
Our main protocol uses 50 generations for all methods. Search is performed on a \texttt{search\_mini} subset to keep evolutionary feedback fast while preserving the fixed holdout and final-test protocol:
\begin{itemize}[leftmargin=1.5em]
    \item \textbf{Fixed, Reasoning, Order-Vote, Order-Vote+}: search\_mini $=256$
    \item \textbf{Order-Rerank}: search\_mini $=128$
\end{itemize}

The asymmetry reflects the much higher inference cost of order-rerank. Importantly, all final comparisons are based on the same selection holdout and independent final test, not on search scores.

We also reevaluate the evolving population against the larger evolution pool at generations 10, 20, 30, 40, and 49. This reduces the risk that a candidate is selected solely because it fits a coarse and noisy search signal. The final generation for each run is chosen on the 665 holdout, and only then is it evaluated on the frozen 855 test examples.

\begin{table}[h]
\centering
\small
\resizebox{\linewidth}{!}{\begin{tabular}{lccc}
\toprule
Comparison & $\Delta$ Final (\%) & 95\% CI & $p$ \\
\midrule
Order-Vote vs. Fixed & 6.55 & [3.27, 9.94] & 0.0000 \\
Order-Vote vs. Order-Rerank & 3.04 & [1.17, 5.03] & 0.0034 \\
\bottomrule
\end{tabular}} 
\caption{Paired bootstrap on representative selected runs for the final test set.}
\label{tab:significance}
\end{table}

\noindent\textbf{Independent repeats.}
ShinkaEvolve does not currently expose a stable global seed interface for fully controlled seed sweeps in our local setup, so we report \emph{independent repeated runs} rather than deterministic seed sweeps. For the three main-table methods we collect five high-budget repeat-only runs under the same protocol. This is the sole source of the mean and standard deviation reported in the main table.

\noindent\textbf{Post-hoc validation.}
For each completed run, we evaluate the selected generation on the final test set and save per-example predictions, modality breakdowns, image-count breakdowns, and paired bootstrap significance results. We also compare each selected generation to its generation-0 ancestor to measure whether evolution itself yields any consistent gains.

Our analysis protocol therefore separates three distinct questions: which strategy looks promising during search, which generation should be selected within a given run, and whether that selected generation actually generalizes to the independent frozen test split. This is especially important for the appendix budget-sensitivity study, where a longer search budget modestly improves holdout while hurting final test.

\subsection{Main Results}

Table~\ref{tab:main-results} is the core empirical result. Under the publication-clean repeat-only high-budget protocol, \textbf{Order-Vote} is the best strategy by a clear margin, reaching $57.89 \pm 0.65\%$ final-test accuracy. This exceeds the stable fixed baseline at $52.73 \pm 0.42\%$ and the stronger but more expensive \textbf{Order-Rerank} variant at $55.79 \pm 0.43\%$.

The effect sizes are large enough to matter. Relative to fixed, order-vote gains 5.17 absolute points on final test. Relative to order-rerank, it gains 2.11 points while also using less post-hoc evaluation budget.

The substantive implication is equally important. The winning method is not the one with the most textual scaffolding, and it is not the one with the deepest second-stage arbitration. It is the strategy whose decision rule is most robust to option-order perturbation. That is why we frame the paper around decision rules rather than prompt length or search depth alone.

Table~\ref{tab:efficiency} shows that the main result is not purchased by brute-force cost. Order-vote requires more calls than the one-shot fixed baseline, but it remains materially cheaper than order-rerank while still outperforming it. This efficiency is critical for MFM-based agents operating in real-world environments, directly addressing the need to reduce computational demands introduced by highly redundant multi-image modalities. On this benchmark, robust aggregation provides the best performance--complexity tradeoff.

\begin{table}[t]
\centering
\small
\resizebox{\linewidth}{!}{\begin{tabular}{lcccc}
\toprule
Method & Holdout Calls & Holdout Sec. & Final Calls & Final Sec. \\
\midrule
Fixed & 665 & 139.0 & 855 & 183.0 \\
Order-Vote & 1330 & 250.0 & 1710 & 331.5 \\
Order-Rerank & 1794 & 328.3 & 2229 & 417.2 \\
\bottomrule
\end{tabular}} 
\caption{Average post-hoc evaluation cost over repeat-only high-budget runs. Order-vote is substantially cheaper than order-rerank while remaining more accurate.}
\label{tab:efficiency}
\end{table}

Table~\ref{tab:weak-baselines} reports the remaining two strategies. \textbf{Reasoning} is useful precisely because it fails in an informative way: it can improve intermediate search behavior without becoming a stronger final method. This makes it a clean negative baseline for the claim that ``more explicit reasoning text'' is enough. \textbf{Order-Vote+} is better viewed as an appendix-only exploration. Although it is occasionally promising, it is not supported by the same repeat-only evidence as the main methods and remains below order-rerank on the latest final-test ranking.

\begin{table}[t]
\centering
\resizebox{\linewidth}{!}{
\begin{tabular}{lcccc}
\toprule
Method & Selected Gen & Gen-0 Final (\%) & Selected Final (\%) & $\Delta$ (\%) \\
\midrule
Fixed & 41 & 52.05 & 52.16 & 0.12 \\
Reasoning & 18 & 51.81 & 51.70 & -0.12 \\
Order-Vote & 21 & 57.66 & 58.71 & 1.05 \\
Order-Rerank & 26 & 55.79 & 55.67 & -0.12 \\
Order-Vote+ & 16 & 53.33 & 55.91 & 2.57 \\
\bottomrule
\end{tabular}} 
\caption{Selected-generation versus generation-0 final-test comparison on representative runs.}
\label{tab:selected-vs-gen0}
\end{table}

Table~\ref{tab:significance} reinforces this ranking with paired bootstrap on representative selected runs. Order-vote significantly outperforms both fixed prompting and order-rerank. The main method choice is therefore supported both by repeated-run averages and by paired item-level comparisons on the final test set. That same Table~\ref{tab:efficiency} clarifies why order-rerank is a \emph{strong ablation} rather than a near-miss winner. It spends substantially more evaluation budget and still does not match the simpler aggregation strategy. On this benchmark, robust aggregation beats heavier arbitration.

\subsection{Analysis}

Figure~\ref{fig:evolution-progress} makes the main dynamic visible without letting noise dominate the narrative. All three main-table methods improve under a 50-generation budget, but the method with the strongest search trajectory is not automatically the one that wins on final test. Fixed improves noticeably in search space yet remains well behind on the final metric. Order-rerank also improves and then largely saturates, but that search behavior does not convert into the strongest final model. Order-vote shows a smaller but cleaner within-budget gain, and it is the only method whose improvement survives the full holdout-to-final pipeline.

\noindent\textbf{Evolution helps, but only modestly.}
Table~\ref{tab:selected-vs-gen0} shows why we keep the paper's claims about evolution deliberately modest. In the representative order-vote run, the selected generation improves over generation 0 by about one final-test point. Fixed changes only marginally, and order-rerank slightly regresses. Evolution therefore matters, but only after the right inference-time strategy has already been chosen.

\noindent\textbf{Breakdown by modality and image count.} Tables~\ref{tab:modality-breakdown} and~\ref{tab:image-count-breakdown} show that order-vote improves over fixed across all reported modalities and image-count buckets. Relative to order-rerank, it is better on CT, MRI, X-ray, and other modalities, and on every image-count bucket in the representative selected runs, with a small ultrasound deficit as the clearest local exception. First, order-vote does not win only because of one dominant modality; it improves over fixed across all reported modalities and all image-count buckets. Second, the advantage over order-rerank is broad rather than fragile: order-vote is better on CT, MRI, X-ray, and other categories, and on every image-count bucket in the representative selected runs, with a small ultrasound deficit as the clearest local exception.

\noindent\textbf{Ultrasound is the clearest local exception.}
In the representative selected-run breakdown, order-rerank slightly outperforms order-vote on ultrasound. We therefore do not claim that order-vote is superior on every slice of the benchmark. Our claim is narrower and more defensible: order-vote is the strongest \emph{overall} strategy and the best performance--complexity tradeoff under the frozen protocol. Appendix Figure~\ref{fig:appendix-case-cards}B makes this concrete with a real final-test item where order-rerank recovers a difficult case that order-vote misses.

\begin{table}[t]
\centering
\small
\begin{tabular}{lccc}
\toprule
Image count & Fixed & Order-Vote & Order-Rerank \\
\midrule
2 images & 56.23 & 58.92 & 56.90 \\
3 images & 48.52 & 55.62 & 55.03 \\
4 images & 45.83 & 65.28 & 62.50 \\
5 images & 51.74 & 58.68 & 53.31 \\
\bottomrule
\end{tabular} 
\caption{Representative final-test breakdown by the number of images in each question.}
\label{tab:image-count-breakdown}
\end{table}

\begin{table}[t]
\centering
\small
\begin{tabular}{lccc}
\toprule
Modality & Fixed & Order-Vote & Order-Rerank \\
\midrule
CT & 55.18 & 62.80 & 60.37 \\
MRI & 50.35 & 53.52 & 49.30 \\
X-ray & 40.91 & 50.00 & 46.59 \\
ultrasound & 55.26 & 61.40 & 64.04 \\
other & 56.10 & 73.17 & 58.54 \\
\bottomrule
\end{tabular} 
\caption{Representative final-test breakdown by imaging modality using the latest complete selected runs for the three methods.}
\label{tab:modality-breakdown}
\end{table}

\noindent\textbf{Why voting beats reranking here.}
Our evidence suggests that this benchmark rewards robust evidence aggregation more than deeper second-stage arbitration. Order-rerank can reach high search scores, but it pays more for each evaluation and exhibits a higher invalid-generation rate. Order-vote instead attacks answer-order sensitivity directly and then aggregates local evidence with a simpler, lower-variance rule. On this task, that bias is better aligned with final generalization than the extra flexibility of reranking.

\noindent\textbf{Budget sensitivity.}
A natural counterargument is that the winning method still has untapped headroom and simply needs more generations. Our evidence argues against that view. In an appendix-only 100-generation ablation, order-vote slightly raises holdout accuracy but \emph{drops} final-test accuracy relative to the 50-generation repeat-only mean. We therefore interpret a longer search as magnifying selection noise rather than improving generalization, and we keep 50 generations as the paper's final budget.

\noindent\textbf{Longer search can overfit the selection signal.}
The 100-generation order-vote ablation slightly improves holdout but hurts the final test. This is exactly the failure mode one would expect when an additional search budget fits a noisy selection signal rather than improving the true target metric.

\noindent\textbf{Search gains do not automatically transfer.}
Reasoning remains the cleanest warning sign. It can look competitive during search or selection without becoming a stronger final-test method. This is precisely why the paper separates search, holdout, and final-test reporting instead of treating search-time improvements as sufficient evidence.

\noindent\textbf{Extra conditional logic is not free.}
Order-Vote+ was intended as a low-risk selective correction layer. In practice, that extra conditional structure does not yet produce a stronger final method than order-rerank, and it is supported by much weaker repeat evidence. We therefore keep it in the appendix rather than promoting it into the main comparison.


\section{Discussion}

The cleanest summary of our findings is that \emph{search is only as good as the agentic decision rule it is searching over}. Reasoning can look better during search without becoming a better final model because its decision rule remains misaligned with the task environment. Order-rerank can search well because it is expressive, yet still lose on final test because that expressiveness creates more ways to overfit intermediate signals. Order-vote wins because it directly targets a central nuisance factor in multiple-choice multi-image reasoning: instability with respect to multimodal signal presentation.

\noindent\textbf{Alignment with Agentic Multimodal Reasoning.}
These findings directly address several core challenges in developing MFM-based agents. When building clinical agents that parse ordered scans, the primary bottleneck is not just the base model's perception, but rather the robustness of its inference-time reasoning policy. By demonstrating that a computationally efficient, order-robust aggregation policy out-generalizes deeper search or heavier deliberative reranking, we highlight a practical path forward for deploying reliable, resource-aware multimodal agents.

This framing also helps explain why the 100-generation ablation is useful even as a negative result. If the dominant bottleneck were insufficient search depth, then a longer run should improve or at least preserve final-test performance. Instead, it slightly improves holdout while harming final test. That directly falsifies the simplest ``just search longer'' explanation.

The broader implication is methodological. In multimodal prompt-program optimization, it is not enough to report that an evolutionary framework can raise an internal search metric. A convincing paper must show which strategy survives a clean holdout-to-final pipeline and report when additional budget does not help \citep{jeong2024medicaladapt,agrawal2025evaluationillusion}. In that sense, ShinkaEvolve matters here less as a standalone endpoint and more as a controlled optimizer that exposes which agentic inference strategies are actually worth searching.


This paper has several important limitations. First, all conclusions are drawn from a single benchmark, MedFrameQA, under a documented internal frozen split rather than an official hidden test server. The protocol is rigorous and reproducible, but it is still an internal split protocol. Second, the study uses one evaluation VLM setup and one local proposal-model setup; stronger claims about cross-model robustness would require repeating the same comparison with additional medical or general-domain backbones, a concern that has surfaced repeatedly in recent medical LVLM evaluations \citep{jeong2024medicaladapt,agrawal2025evaluationillusion}. Third, our repeat protocol is based on independent repeated runs rather than deterministic seed sweeps because the local ShinkaEvolve stack does not currently expose a stable end-to-end global seed for the full pipeline.

Our study is a systems-and-evaluation paper, not a clinical deployment study. The methods are not intended for direct diagnostic use, and we do not claim readiness for real clinical decision support. All quantitative gains should be interpreted as benchmark improvements under a frozen experimental protocol, not as evidence of safety, fairness, or deployability in healthcare settings.

Finally, our budget-sensitivity findings should be interpreted at the level of this benchmark and this method class. The negative 100-generation result is strong evidence that longer search is unnecessary in our setting, but it does not imply that larger search budgets are always harmful for every prompt-evolution task. Rather, it shows that longer search should not be assumed to help without final-test validation.

\section{Conclusion}

We presented a controlled study of prompt-program evolution for multi-image medical VQA on MedFrameQA. Under a fixed high-budget ShinkaEvolve protocol and a reproducible internal frozen split, the main conclusion is sharp: {inference-time agentic decision rules matter more than additional search budget}. A lightweight order-vote policy consistently outperforms both a stable reactive baseline and a stronger but more computationally expensive order-rerank variant. Extending the search from 50 to 100 generations does not improve final generalization and can even degrade it somewhat. More broadly, the paper argues for a stricter way of studying prompt-program evolution in multimodal settings: one must ask what underlying agentic policy is actually being optimized, how candidates are selected, and whether larger budgets improve the true reporting metric rather than only an intermediate signal.


\bibliographystyle{acl_natbib}
\bibliography{references}

\newpage
\appendix
\section{Appendix}

\subsection{Budget Sensitivity: 50 vs.\ 100 Generations}

\begin{table}[h]
\centering
\small
\resizebox{\linewidth}{!}{\begin{tabular}{lccc}
\toprule
Budget setting & Holdout (\%) & Final (\%) & Invalid (\%) \\
\midrule
Order-Vote 50-gen mean & 62.50 & 57.89 & --- \\
Order-Vote 100-gen single & 62.86 & 56.02 & 1.00 \\
\bottomrule
\end{tabular}}
\caption{Budget sensitivity for Order-Vote. Relative to the 50-generation repeat-only mean, the 100-generation run changes holdout by 0.36 points and final test by -1.87 points.}
\label{tab:budget-sensitivity}
\end{table}

\begin{figure}[h]
    \centering
    \includegraphics[width=0.75\linewidth]{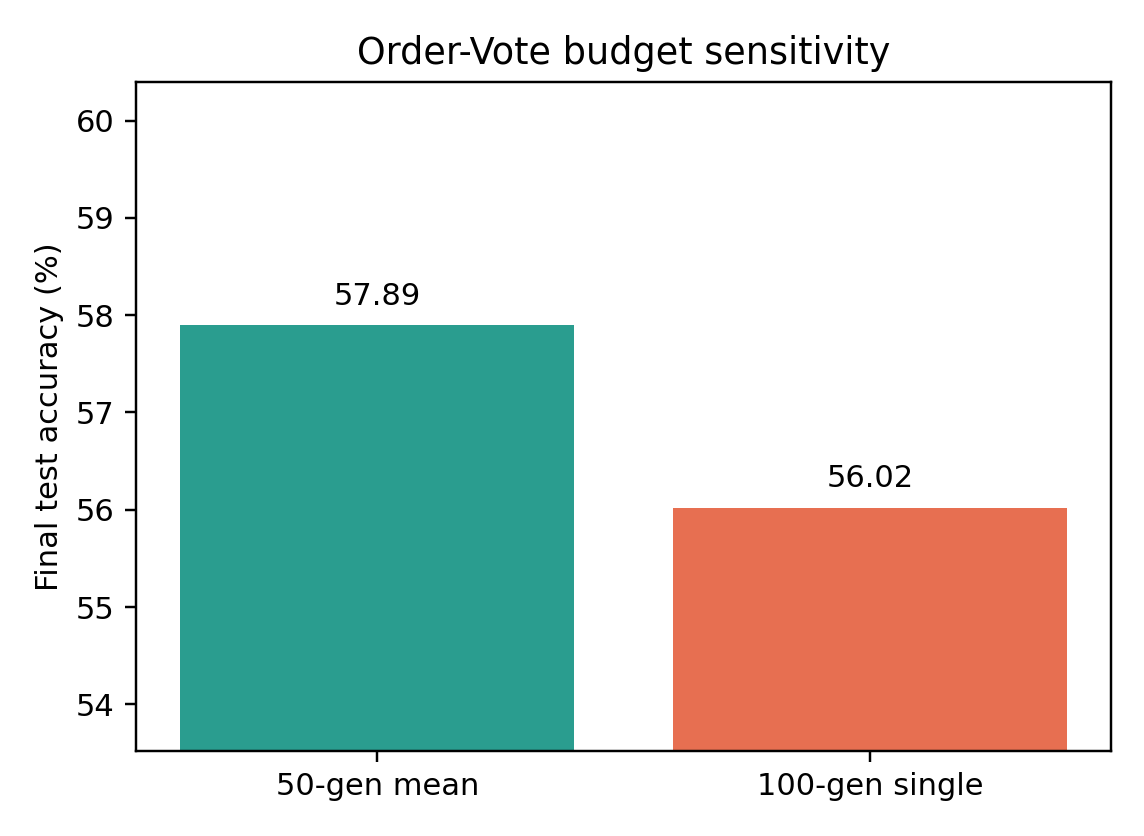}
    \caption{Final-test comparison between the 50-generation repeat-only mean and the single 100-generation order-vote ablation. The longer search budget does not improve final performance.}
    \label{fig:budget-sensitivity}
\end{figure}

The 100-generation order-vote run reaches a slightly higher holdout score than the 50-generation repeat-only mean, but its final-test score drops by 1.87 absolute points. We therefore keep 50 generations in the main paper and treat the 100-generation result as a negative budget-sensitivity finding.

\subsection{Representative Evolution Gains over Generation 0}

These results reinforce the main paper's interpretation: evolution contributes a modest gain for order-vote, little change for fixed, and no reliable gain for the more complex reranking strategy. This is why we frame evolution as a \emph{secondary} factor: it helps after the right strategy has been chosen, but it does not substitute for the right inference-time structure.

\begin{figure*}[h]
    \centering
    \includegraphics[width=0.82\textwidth]{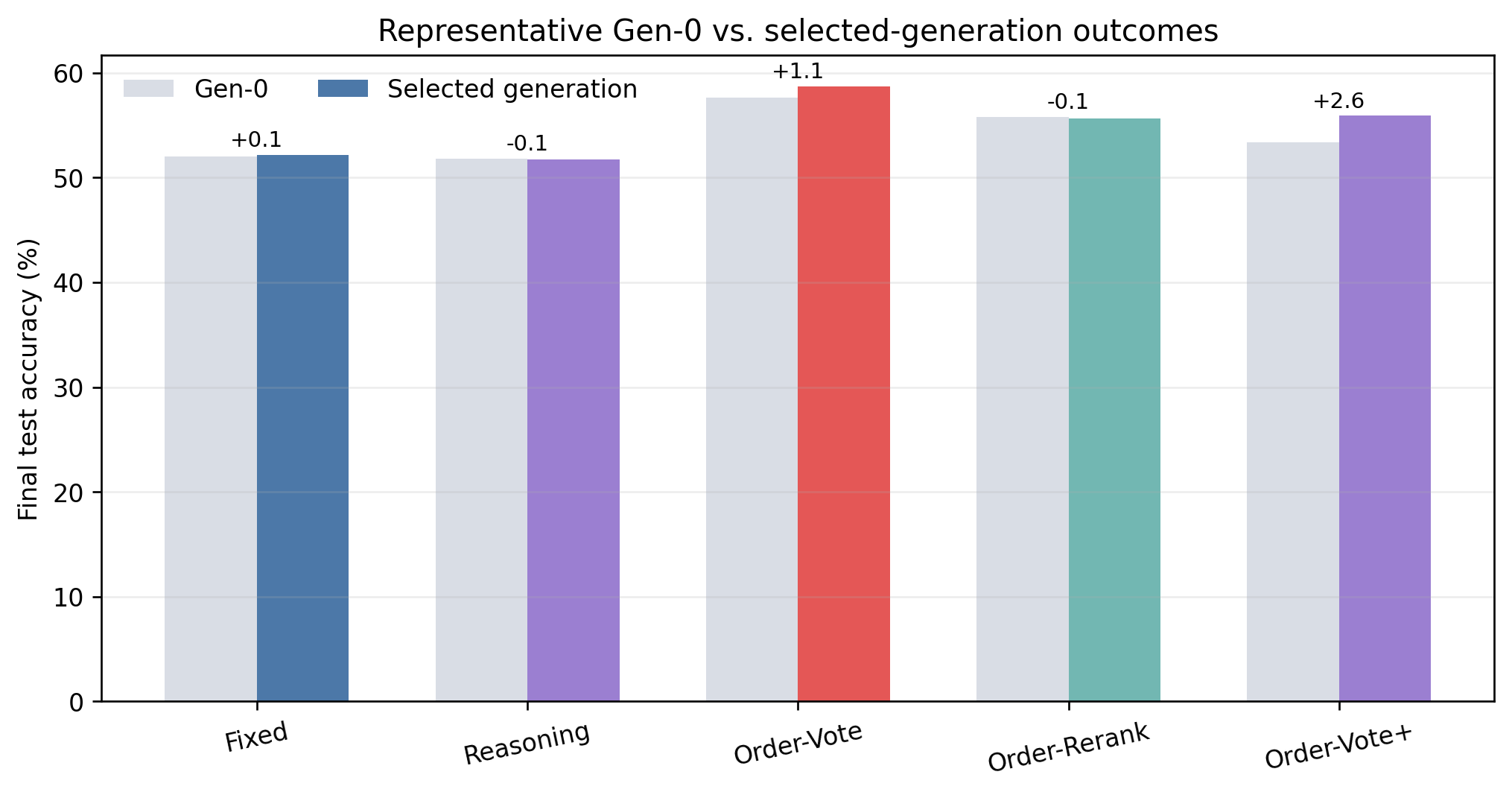}
    \caption{Representative final-test change from generation 0 to the selected generation. The strongest final method is not the one with the largest raw search movement; what matters is whether evolution produces a selected program that survives final-test validation.}
    \label{fig:appendix-selected-vs-gen0}
\end{figure*}

\subsection{Raw Search Dynamics and Selection Mismatch}

\begin{figure*}[h]
    \centering
    \includegraphics[width=0.95\textwidth]{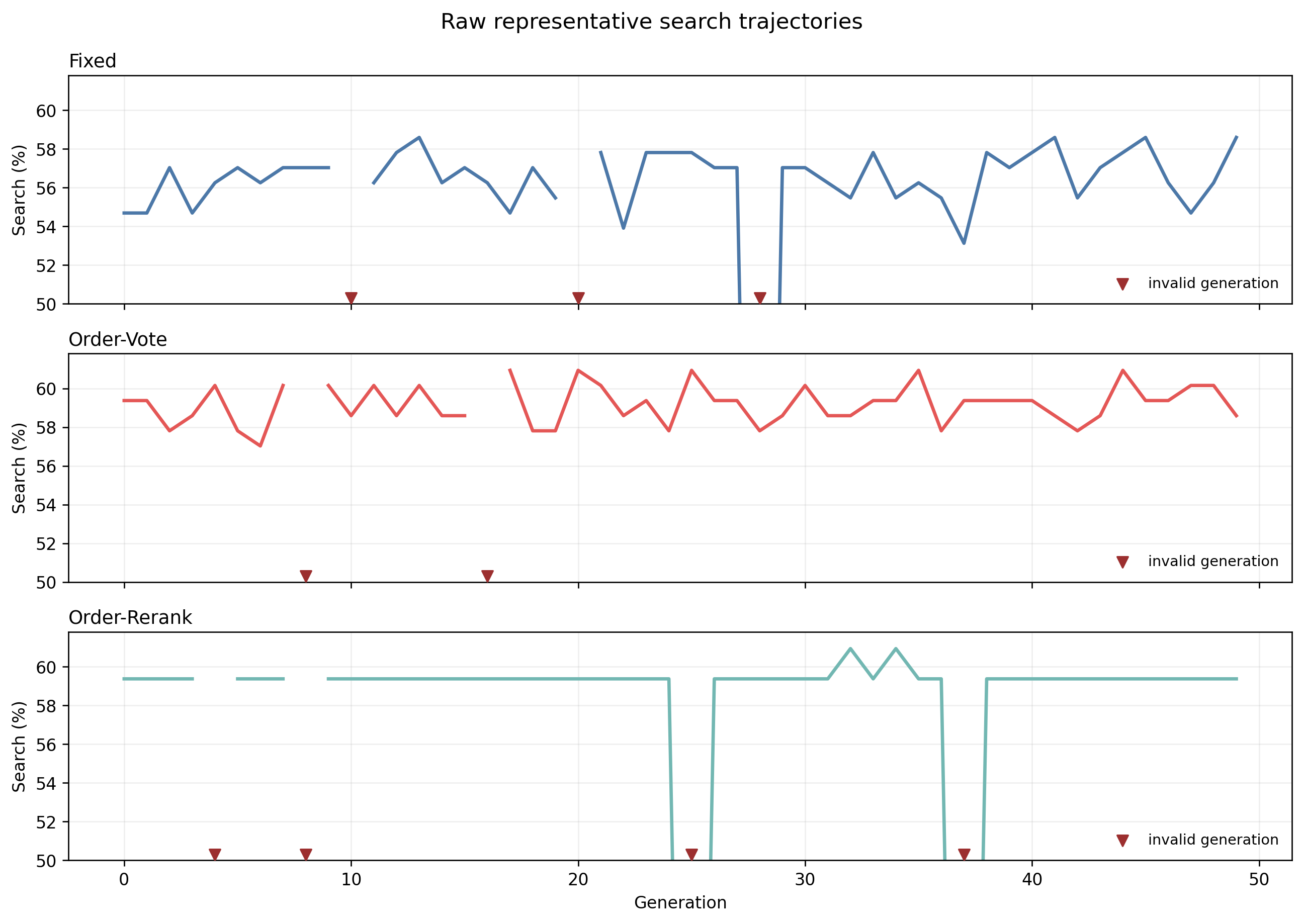}
    \caption{Raw representative search trajectories for the three main-table methods. The appendix keeps the noisy view, including invalid generations, while the main text uses best-so-far progress to show within-budget improvement more clearly.}
    \label{fig:appendix-raw-evolution}
\end{figure*}

\begin{figure}[h]
    \centering
    \includegraphics[width=\linewidth]{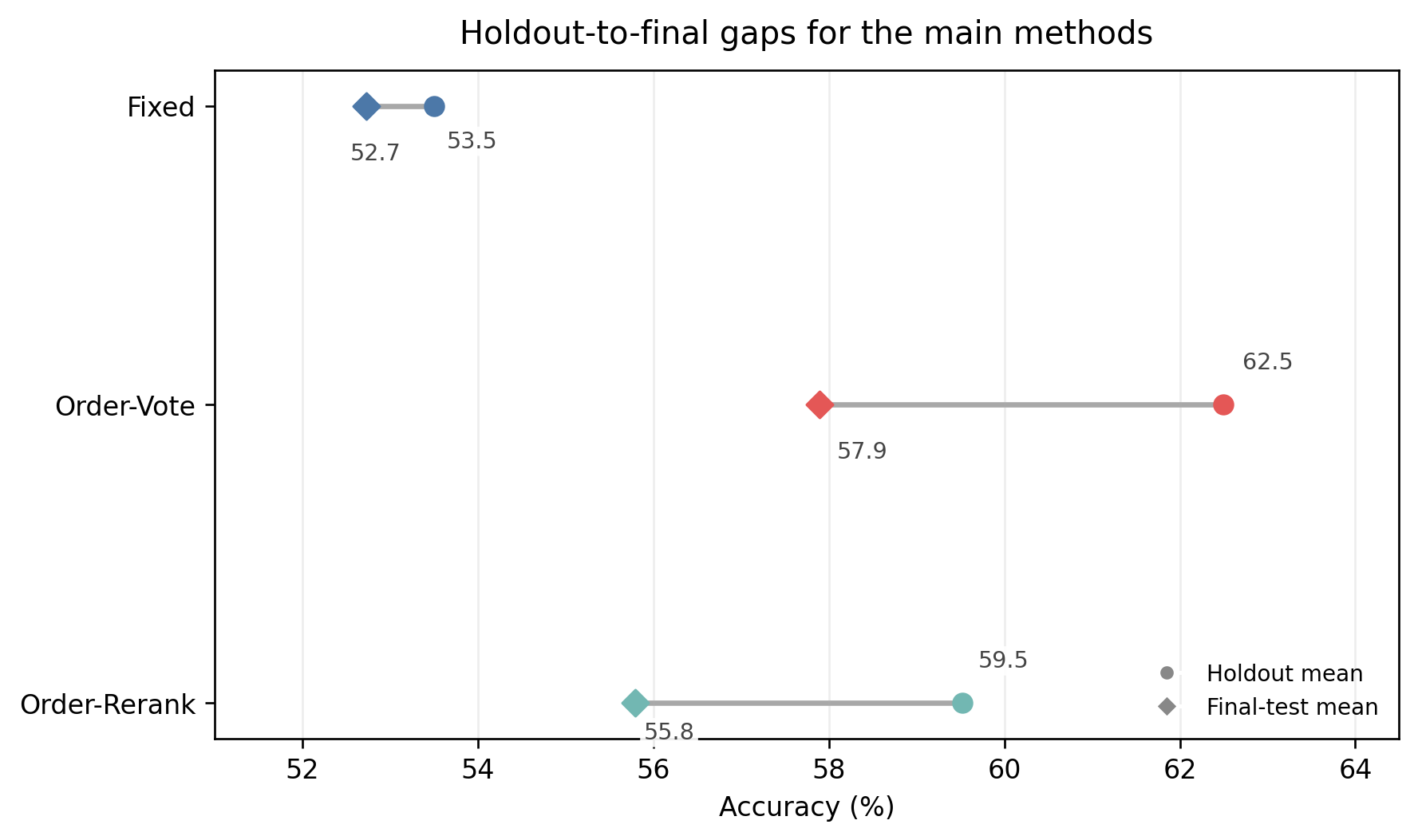}
    \caption{Selection-holdout versus final-test means for the three main-table methods. All three methods lose accuracy when moving from holdout to final test, which is why the paper treats search and selection signals as insufficient proxies for true generalization.}
    \label{fig:appendix-selection-gap}
\end{figure}

Appendix Figure~\ref{fig:appendix-raw-evolution} shows the underlying search traces that sit behind the cleaner main-text progress figure. The raw curve s contain local noise and a small number of invalid generations, especially for the more complex reranking strategy, but they still show that all three methods receive useful search-time feedback under the 50-generation budget. Appendix Figure~\ref{fig:appendix-selection-gap} then closes the loop by showing why the paper does not rank methods on holdout alone: even strong search and holdout behavior can overstate final-test quality, and that gap is largest for the more fragile methods.

\subsection{Failure Cases and Boundary Conditions}

Our conclusions are strong, but they are not universal in a naive sense. We therefore make the main boundary conditions explicit. Appendix Figure~\ref{fig:appendix-case-cards}A--C provides one positive case, one boundary case, and one failure case to anchor these statements qualitatively.




\subsection{Prompt-Space Design}

The prompt spaces used in this study are intentionally restricted. ShinkaEvolve does not rewrite arbitrary runtime logic. Instead, each strategy exposes a method-specific configuration block while data loading, split handling, evaluation, and post-hoc validation remain frozen.

\begin{table*}[h]
\centering
\small
\begin{tabular}{p{1.7cm}p{5.2cm}p{7.0cm}}
\toprule
Strategy & Evolvable configuration space & Frozen outer behavior \\
\midrule
Fixed & Short instruction wording, image framing phrasing, decision wording, answer-format cues & All images are presented once in original order and the model returns a single answer letter \\
Reasoning & Case role, frame-inspection strategy, sequence strategy, option comparison wording, final decision-rule text & Still a single-pass all-image decision; only the reasoning scaffold changes \\
Order-Vote & Number of reordered views, vote tie-break rule, local-view wording, fallback behavior & Local predictions are mapped back to the original label space and aggregated by vote \\
Order-Rerank & Number of reordered views, rerank top-k policy, pairwise comparison wording, fallback behavior & Voting is followed by a conditional second-stage rerank among top candidates \\
Order-Vote+ & Uncertainty trigger rule, top-2 correction wording, fallback behavior & A second-stage correction is only allowed when the first-stage vote is uncertain \\
\bottomrule
\end{tabular}
\caption{Prompt-space design. The evolvable content differs by strategy, but the surrounding runtime and evaluation protocol are frozen.}
\label{tab:prompt-space}
\end{table*}

This restriction is central to the paper's logic. The goal is not to discover arbitrary task-specific Python programs. It is to compare interpretable agentic inference strategies under the same search engine and the same validation protocol.

\subsection{Qualitative Example and Real Case Cards}

Figure~\ref{fig:task-example} is included primarily as a task-format illustration rather than as a claim-bearing case study. We use it in the supplement for a simple reason: it makes visible the type of ambiguity the benchmark is built around. The question refers to a later-stage CT image, so the answer cannot be inferred from one frame alone. This is representative of the benchmark's core challenge and helps explain why option-order robustness and cross-frame evidence aggregation become more important than simply adding more reasoning text.

We deliberately avoid overclaiming from a single qualitative example. The paper's central conclusions are supported by repeated-run final-test evidence, not by isolated cherry-picked cases. Still, examples like Figure~\ref{fig:task-example} are useful because they make the benchmark's multi-image structure concrete and help motivate why a strategy such as order-vote can be a better inductive match than either a fixed prompt or a heavier reranking stack.

\begin{figure*}[t!]
    \centering
    \includegraphics[width=\textwidth]{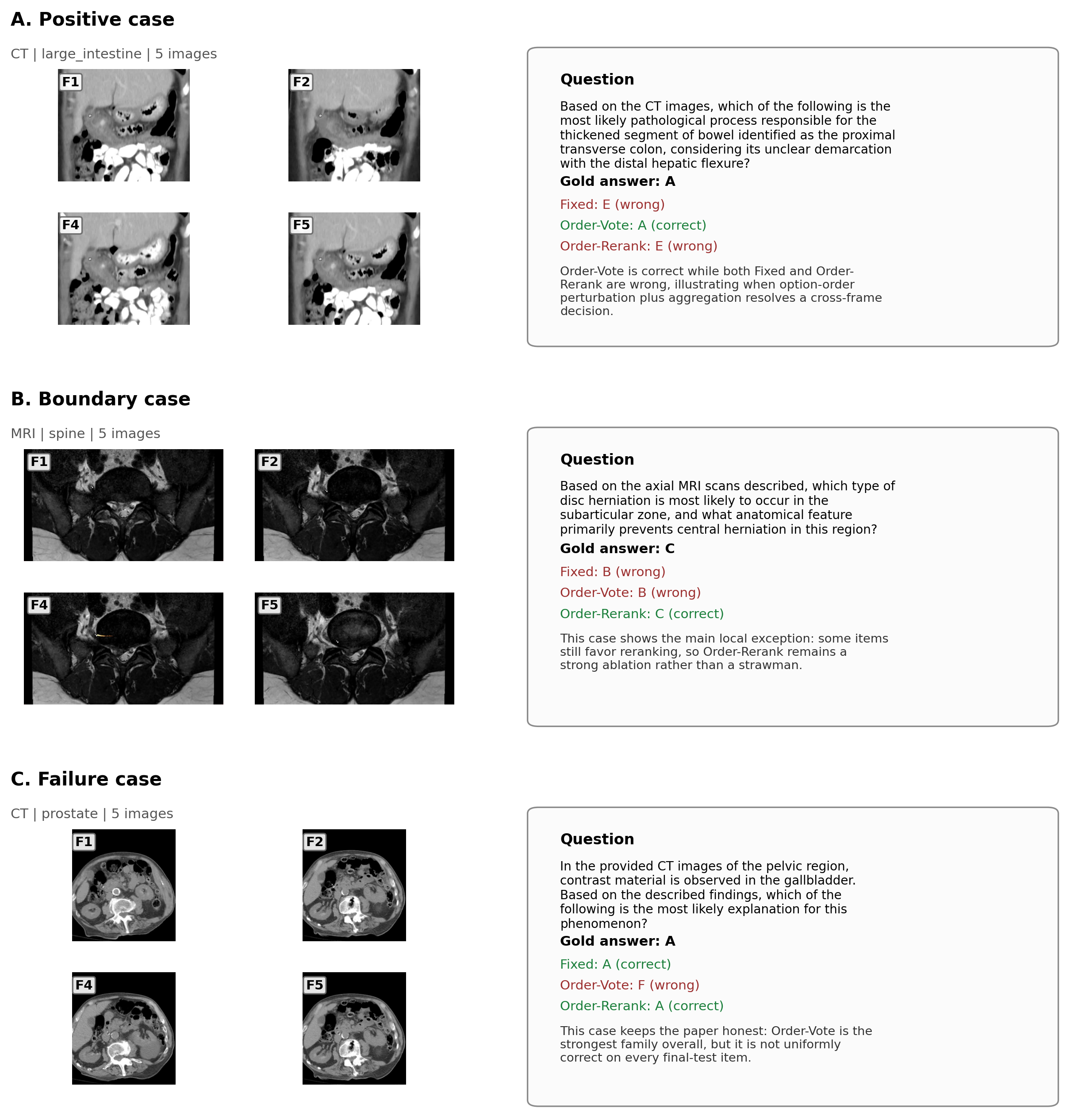}
    \caption{Three case cards drawn from representative selected-run final-test predictions. Case A is a positive example where Order-Vote is correct, while both Fixed and Order-Rerank are wrong. Case B is a boundary example where Order-Rerank recovers a difficult item that Order-Vote misses. Case C is a failure example showing that Order-Vote is the strongest strategy overall, but not universally correct on every item.}
    \label{fig:appendix-case-cards}
\end{figure*}

To complement that task-format illustration, Figure~\ref{fig:appendix-case-cards} shows three case cards mined directly from the representative selected-run final-test predictions used throughout the paper. These are not new experiments, and they are not substitutes for the repeated-run quantitative evidence. Instead, they serve a narrower purpose: they show what our claims look like at the item level when the aggregate findings are instantiated on real benchmark examples.

Case A supports the main paper thesis at the item level: when the answer depends on aggregating evidence across multiple frames, order-vote can resolve a decision that both the fixed one-shot baseline and the heavier reranking strategy tend to miss. Case B supports the narrower claim we make about boundary conditions: order-rerank is a real competitor and remains worth reporting because it can still win on certain hard items, especially in the local regime where our modality breakdown already shows a smaller gap. Case C keeps the narrative honest by showing that order-vote does not dominate every single question; the paper's claim is about overall repeated-run generalization, not universal per-item superiority.

\subsection{Implementation Notes and Reproducibility}

All experiments use the same internal frozen split with sizes 1331/665/855 for evolution pool, selection holdout, and independent final test respectively. The split is group-preserving by \texttt{video\_id} and versioned as \texttt{medframeqa\_split\_manifest\_v2}. Prompt-program evolution uses the same high-budget ShinkaEvolve profile across the main methods, with the only search-stage asymmetry being a smaller search\_mini for order-rerank because of its substantially higher inference cost.

\paragraph{Search profile.}
The high-budget profile used for the main paper consists of 5 islands, archive size 40, weighted parent selection, migration every 10 generations with migration rate 0.1, and up to 3 patch attempts per generation. We enable textual feedback to the proposal model and restrict patch styles to \texttt{diff} and \texttt{full} edits because cross-style edits were empirically too destructive for our configuration-like prompt spaces.

\paragraph{What is frozen and what evolves.}
The runtime, evaluation, split, and post-hoc validation code are frozen. What evolves is the restricted method-specific configuration block that determines the strategy's prompt wording and lightweight decision settings. This restriction is important for interpretability: we are not evolving arbitrary Python logic, but structured prompt programs with fixed outer behavior.

\paragraph{Why Order-Vote+ stays in the appendix.}
Order-Vote+ is intentionally reported as an exploratory appendix variant rather than promoted into the main comparison. Although it can help on some runs, it is currently supported by only one complete repeat-only run and does not clear the stronger order-rerank baseline on final test. We therefore keep it as a useful negative-or-mixed finding instead of overclaiming a new main method.

\end{document}